# A Self-Supervised Learning-based Approach to Clustering Multivariate Time-Series Data with Missing Values (SLAC-Time): An Application to TBI Phenotyping

Hamid Ghaderi [a, *], Brandon Foreman [b], Amin Nayebi [a], Sindhu Tipirneni [c], Chandan K. Reddy [c], Vignesh Subbian [a, d]

[a] Department of Systems and Industrial Engineering, University of Arizona, Tucson, AZ, USA

[b] College of Medicine, University of Cincinnati, Cincinnati, OH, USA

[c] Department of Computer Science, Virginia Tech, Arlington, VA, USA

[d] Department of Biomedical Engineering, University of Arizona, Tucson, AZ, USA

**Abstract**

Self-supervised learning approaches provide a promising direction for clustering multivariate time-series data. However, real-world time-series data often include missing values, and the existing approaches require imputing missing values before clustering, which may cause extensive computations and noise and result in invalid interpretations. To address these challenges, we present a **S**elf-supervised **L**earning-based **A**pproach to **C**lustering multivariate **T**ime-series data with missing values (SLAC-Time). SLAC-Time is a Transformer-based clustering method that uses time-series forecasting as a proxy task for leveraging unlabeled data and learning more robust time-series representations. This method jointly learns the neural network parameters and the cluster assignments of the learned representations. It iteratively clusters the learned representations with the K-means method and then utilizes the subsequent cluster assignments as pseudo-labels to update the model parameters. To evaluate our proposed approach, we applied it to clustering and phenotyping Traumatic Brain Injury (TBI) patients in the Transforming Research and Clinical Knowledge in Traumatic Brain Injury (TRACK-TBI) study. Clinical data associated with TBI patients are often measured over time and represented as time-series variables characterized by missing values and irregular time intervals. Our experiments demonstrate that SLAC-Time outperforms the baseline K-means clustering algorithm in terms of silhouette coefficient, Calinski Harabasz index, Dunn index, and Davies Bouldin index. We identified three TBI phenotypes that are distinct from one another in terms of clinically significant variables as well as clinical outcomes, including the Extended Glasgow Outcome Scale (GOSE) score, Intensive Care Unit (ICU) length of stay, and mortality rate. The experiments show that the TBI phenotypes identified by SLAC-Time can be potentially used for developing targeted clinical trials and therapeutic strategies.

---

* Corresponding author: ghaderi@arizona.edu

**1. Introduction**

Multivariate time-series data are frequently observed in many healthcare domains where each patient is represented by a set of clinical measurements recorded over time and present important information spanning the whole course of a patient's care. Clustering approaches are commonly used to extract valuable information and patterns from multivariate time-series data [1]. Such clustering approaches can be broadly divided into two categories: raw data-based approaches and representation-based approaches [2]. Raw data-based approaches perform the clustering on raw input data using well-designed similarity measures that can address the specificities of the temporal dimension, including shifted or stretched patterns (e.g., [3–5]). However, since all time points are included, raw data-based clustering approaches are highly susceptible to noise and outliers [2]. Representation-based clustering approaches, on the other hand, employ clustering methods on the representations learned from input time series, which mitigates the effects of noise and outliers in raw input data [2]. Representation learning techniques aim to eliminate the time dimension while preserving the relationship between nearby data points, or they aim to make the comparison more accurate by aligning time-series data with each other [6]. Deep learning architectures have strong representation-learning capabilities, making them useful for state-of-the-art supervised and unsupervised methods to learn the representations of time series for different downstream tasks [7]. Non-linear mappings can be learned using deep learning, allowing time-series data to be transformed into representations that are more suited for clustering [7]. However, deep learning models need to be trained via supervised learning that requires a large, annotated dataset [8]. Building such a dataset would be too costly or often not feasible in most clinical applications. Considering this, self-supervised learning is becoming more common for clustering multivariate time-series data [8], where a deep learning model is trained on an unlabeled time-series dataset by performing a proxy task, and then the learned representations are applied to the clustering task.

*Self-supervision and missingness:* Existing self-supervised learning-based methods for clustering multivariate time-series data are only effective in scenarios with no missing values, while multivariate time-series data are rarely complete due to a variety of reasons. There are three ways to address the missing values in multivariate time-series data: (1) omitting entire samples including missing data, (2) filling in the missing data using data imputation or interpolation methods, and (3) aggregating the irregularly sampled data into discrete time periods. Omitting the samples with missing data and performing the analysis just on the available observations is a straightforward method, but it does not perform well when the rate of missingness is high and/or when the samples are insufficient [6]. Data imputation is another solution that involves substituting new values for the ones that are missing. However, imputing missing values in multivariate time-series data without having domain knowledge about each time-series variable can lead to bias and invalid conclusions. Interpolation techniques are straightforward and commonly used in real-world settings to address missing values. These techniques, meanwhile, may not be able to capture complicated patterns of multivariate time-series data since they do not consider correlated variables [6]. Additionally, when time-series data are sparser, interpolation methods often degrade by adding unwanted noise and additional complexity to the model. Another issue associated with multivariate time-series data is that they may include different time-series variables measured at irregular time intervals. Aggregating measurements into discrete time periods is a typical strategy for addressing irregular time intervals, but this method results in loss of granular information [9]. To address these challenges, we propose a novel **S**elf-supervised **L**earning-based **A**pproach to **C**lustering multivariate **T**ime-series data with missing values (SLAC-Time) that does not rely on any data imputation or aggregation methods. We evaluate the proposed approach by applying it to the problem of clustering time-series data collected from acute traumatic brain injury (TBI) patients. TBI patients exhibit considerable variability in their clinical presentation, making it challenging to identify effective interventions [10,11]. However, by leveraging an advanced clustering technique, TBI patients can be stratified into distinct phenotypic groups with greater precision and reliability that would allow for targeted interventions or clinical studies [10]. **Figure 1** shows a high-level overview of our work that addresses this problem.

Motivated by the shortcomings of the existing state-of-the-art clustering approaches, in this work, we make the following contributions:

- We propose a novel self-supervised learning-based clustering approach called SLAC-Time for clustering multivariate time-series data with missing values without using any data imputation or aggregation methods.
- We perform time-series forecasting as a proxy task for learning more robust representations of unlabeled multivariate time-series data.
- We demonstrate the ability of SLAC-Time in identifying reliable TBI patient phenotypes and their distinct baseline feature profiles using TBI clinicians' domain knowledge and different cluster validation methods.

The rest of this paper is organized as follows. In section 2, we review relevant work in TBI clustering and deep learning-based clustering of multivariate time-series data. Section 3 presents the problem formulation and provides a detailed description of SLAC-Time. Section 4 presents the implementation of SLAC-Time to cluster TBI patients, along with the validation of identified phenotypes using different internal and external validation methods. Finally, Section 5 concludes our work and suggests future directions for research.

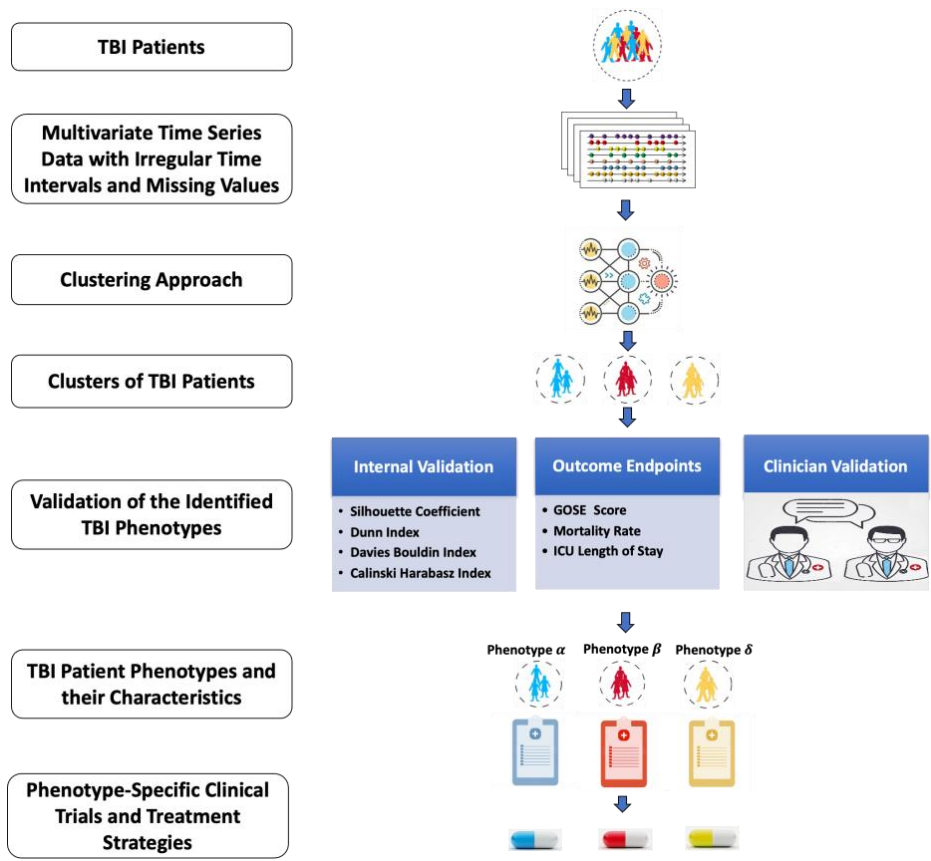

Figure 1. A high-level overview of the problem of identifying TBI phenotypes. TBI: Traumatic Brain Injury; GOSE: Glasgow Outcome Scale-Extended; ICU: Intensive Care Unit

## 2. Related work

In this section, we review existing methods for clustering TBI patients as well as state-of-the-art self-supervised learning-based approaches to clustering multivariate time-series data.

### 2.1. Clustering TBI patients

Existing methods for clustering TBI patients are limited to using non-temporal features with a need for imputing missing values. Folweiler et al. [10] implemented a wrapper framework consisting of two stages. In the first stage, generalized low-rank models were used for selecting significant TBI variables. Then, in the second stage, the selected variables were used for clustering TBI patients into distinct phenotypes by a partitional clustering method. Multivariate imputation by chained equations (MICE) was used with the random forest method to impute missing values.

Si et al. [12] used a sparse hierarchical clustering method for subgrouping patients with mild TBI while using the MICE method to impute the missing values of the features. Yeboah et al. [13] proposed a framework for an explainable ensemble clustering model, including K-means, spectral, Gaussian, mixture, and agglomerative clustering methods to identify TBI phenotypes. They excluded patients with missing data among key features and only included those records with less than 1% missing values and used imputation by mean. As one of the first efforts to cluster TBI patients without imputing missing values of the features, Akerlund et al. [14] developed an unsupervised learning method based on probabilistic graph models of TBI patients' early clinical and laboratory data. Despite clustering without data imputation methods, their cluster analysis approach does not incorporate time-series features that are commonly measured in the Intensive Care Unit (ICU).

## 2.2. Self-supervised learning-based approaches for clustering multivariate time-series data

Self-supervised learning-based approaches often include two stages for clustering multivariate time-series data: (1) learning feature vectors or representations of multivariate time-series data and (2) clustering the learned representations. These methods first convert the input multivariate time-series data into low-dimensional representations; then, the clustering techniques are applied to the learned representations. Tavakoli et al. [15] proposed a two-stage autoencoder-based approach to cluster time-series data with no labels and features. In the first stage of their proposed approach, descriptive metadata were captured as features. Subsequently, K-means clustering method was applied to the extracted features to identify their cluster labels which are then utilized as the labels of input time series. Although this approach clusters time-series data based on their known and hidden non-linear features, it does not handle missingness in time-series data. In another work, Ma et al. [16] proposed a self-supervised time-series clustering network (STCN), a clustering approach that simultaneously optimizes representation learning and clustering. In the representation learning module of this approach, a recurrent neural network (RNN) performed a one-step time-series prediction, and the parameters of the output layer were regarded as model-based representations. Then, these representations were supplied into a self-supervised learning module in which a linear classifier obtains pseudo-labels initialized by K-means. Furthermore, spectral analysis was performed to limit comparable representations to have the same pseudo-labels and match the predicted labels with pseudo-labels. Due to the lack of a strategy for discovering the correlation between the time-series variables, STCN can only be used for clustering univariate time-series data. Moreover, this method does not address missingness and irregular time intervals in time-series data, making it inappropriate for clustering clinical data that usually include multivariate time-series data with missing values. To address issues with clustering clinical data, Jong et al. [17] proposed a variational deep embedding with recurrence (VaDER) approach based on an extended Gaussian mixture variational autoencoder for clustering clinical multivariate time-series data with missing values. However, to handle missing values, they integrated a data imputation scheme into model training, which can result in unnecessary computations and noise.

Real-world time-series data often include missing values which can be an issue, especially in clinical data analysis [18]. Our literature review shows that related clustering approaches learn the representation of time series either when there is no missing value or when the missing values are imputed beforehand. To the best of our knowledge, there is no self-supervised learning-based clustering approach that handles missing values in multivariate time-series data without resorting to imputation methods.

## 3. Methods
### 3.1. Preliminaries

Representation-based clustering approaches necessitate the use of an effective representation learning technique that best suits the type of input data. SLAC-Time leverages a self-supervised Transformer model for time series (STraTS) to learn the representations of multivariate time-series data. It maps input multivariate time-series data with missing values into a fixed-dimensional vector space without resorting to data imputation or aggregation methods [9]. Therefore, unlike traditional methods which treat each multivariate time-series data as a matrix with specific dimensions, our approach treats each multivariate time-series data as a set of observation triplets, avoiding the need for data imputation or aggregation. The Transformer-based architecture of STraTS uses self-attention to go from one token to another in a single step, allowing for parallel processing of observation triplets. Observation triplets are embedded using a novel Continuous Value Embedding (CVE) method, eliminating the necessity for binning continuous values before embedding them. By doing so, the fine-grained information that is lost when time is discretized is preserved. We denote STraTS mapping by $f_\theta$ with $\theta$ being the model's parameters. We use the term "representation" to refer to the vector that results from applying this mapping to a multivariate time-series data.

Given a training set of $N$ unlabeled samples represented by $\mathcal{D} = \{(d^k, T^k)\}_{k=1}^N$ where the $k^{th}$ sample includes a non-temporal vector $d^k \in \mathbb{R}^D$ and a multivariate time-series data $T^k$, we determine the optimal parameter $\theta^*$ such that the mapping $f_{\theta^*}$ yields general-purpose representations. To do so, drawing from the STraTS model, we define each time-series feature as a set of observation triplets where each triplet is of the form $(t, f, v)$ where $t \in \mathbb{R}_{\geq 0}$ is the time of measurement, $f$ is the name of the variable, and $v \in \mathbb{R}$ represents the value of the variable. A multivariate time-series data $T$ of length $n$ is defined as a set of $n$ observation triplets where $T = \{(t_i, f_i, v_i)\}_{i=1}^n$. To obtain the representation of samples, we use the STraTS model from our prior work in [9]. With an input multivariate time-series data $T = \{(t_i, f_i, v_i)\}_{i=1}^n$, the initial embedding for the $i^{th}$ triplet $e_i \in \mathbb{R}^d$ is calculated by adding the three embeddings, including (1) feature embedding $e_i^f \in \mathbb{R}^d$, (2) value embedding $e_i^v \in \mathbb{R}^d$, and (3) time embedding $e_i^t \in \mathbb{R}^d$ [9]. To put it another way, $e_i = e_i^f + e_i^v + e_i^t \in \mathbb{R}^d$. Feature embeddings, such as word embeddings, are produced using a basic lookup table. Value embeddings and time embeddings are obtained by a one-to-many Feed-Forward Network (FFN) as follows [9]:

$$e_i^v = FFN^v(v_i) \tag{1}$$

$$e_i^t = FFN^t(t_i) \tag{2}$$

Both FFNs consist of one input neuron, $d$ output neurons, a single hidden layer containing $\lfloor \sqrt{d} \rfloor$ neurons and a $\tanh(.)$ activation function. These networks can be represented as

$$FFN(x) = U \tanh(Wx + b) \tag{3}$$

where the dimensions of the weight parameters $\{W, b, U\}$ are determined by the sizes of the hidden and output layers within the FFN. The initial embeddings $\{e_1, \ldots, e_n\} \in \mathbb{R}^d$ are processed by a Transformer [19] consisting of $M$ blocks. Each block contains a Multi-Head Attention ($MHA$) layer with h attention heads and an FFN with one hidden layer. Each block's output serves as the input for the subsequent block, with the final block's output yielding the contextual triplet embeddings $\{c_1, \ldots, c_n\} \in \mathbb{R}^d$. Then, a self-attention layer is utilized to compute time-series embedding $e^T \in \mathbb{R}^d$ [9]. We also obtain the embedding of non-temporal variables by passing $d$ through an FFN [9].

**3.2. SLAC-Time architecture**

The architecture of SLAC-Time is illustrated in Figure 2. SLAC-Time defines its input as a set of observation triplets that pass through STraTS model to generate the representation. Considering an unlabeled input dataset, SLAC-Time generates pseudo-labels to fine-tune the self-supervised model and facilitate performing the target task. In the target task, pseudo-labels serve as target classes for unlabeled data as though they were actual labels. SLAC-Time contains three modules, including (1) self-supervision, (2) pseudo-label extraction, and (3) classification module. This approach alternates between the pseudo-label extraction and classification modules to update cluster assignments and increase the quality of clusters. The following is a detailed description of SLAC-Time architecture.

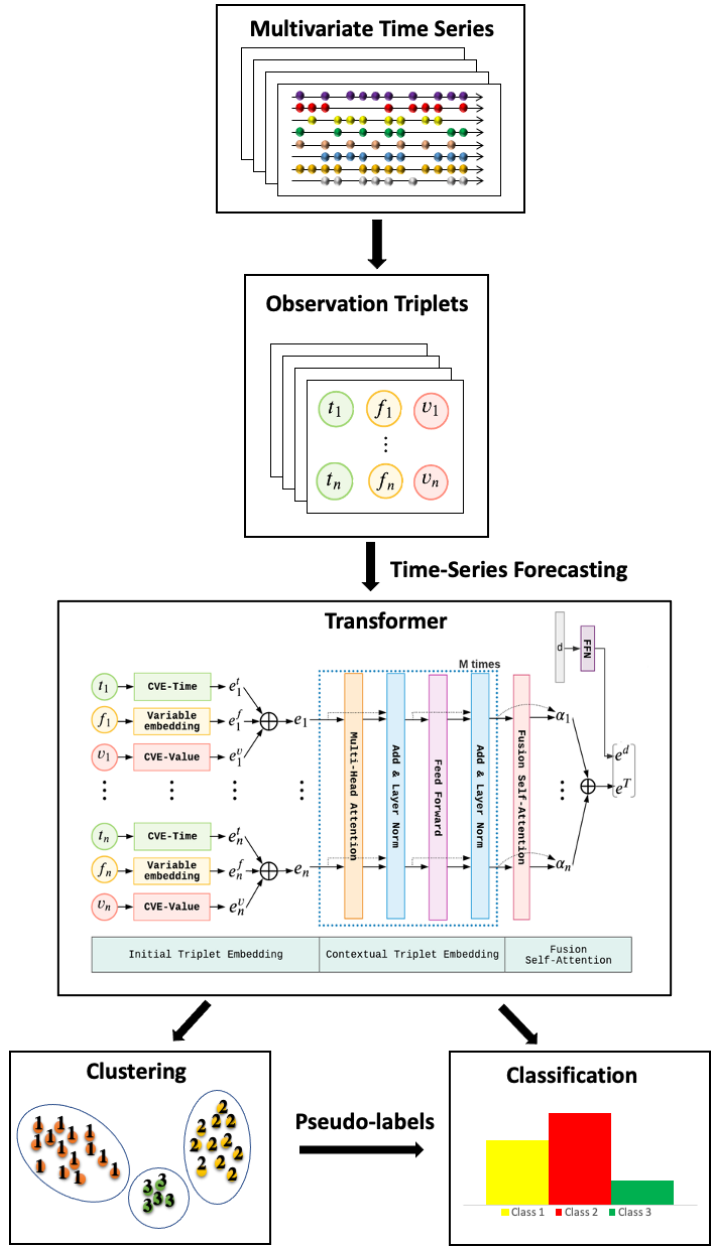

Figure 2. An overview of the proposed SLAC-time clustering approach

### 3.2.1. Self-supervision module

In the first step of SLAC-Time, we pre-train STraTS by performing time-series forecasting as a self-supervision task to learn the representation of the unlabeled multivariate time-series data. To do so, we use a larger dataset with $N' \geq N$ samples given by $\mathcal{D}' = \{(d^k, T^k, m^k, z^k)\}_{k=1}^{N'}$, where $m^k \in \mathbb{R}$ represents the forecast mask, indicating whether a variable was observed in the forecast window, and $z^k \in \mathbb{R}$ includes the associated values of the variable. We need to mask out the unobserved forecasts in the loss function because they cannot be used for training the model. The time-series data in the forecast task dataset are created by considering various observation windows in the time series. **Figure 3** depicts how we construct inputs and outputs for the forecast task.

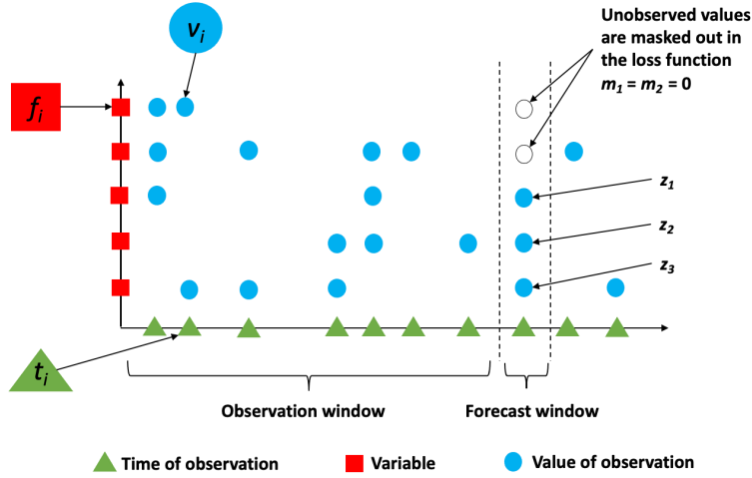

Figure 3. An illustration of input and output in the forecasting task

The time-series forecasting output is obtained by passing the concatenation of non-temporal and time-series embeddings through the following layer [9]:

$$\tilde{z} = W_s [e^d \ e^T] + b_s \in \mathbb{R}^{|\mathcal{F}|} \tag{4}$$

where $W_s$ and $b_s$ are the weights, and $e^T$ and $e^d$ represent the time-series embedding and the non-temporal embedding, respectively.

To address missing values in the forecast outputs, we use a masked Mean Squared Error (MSE) loss for training on the forecast task model. The self-supervision loss is defined as

$$\mathcal{L}_{ss} = \frac{1}{|N'|} \sum_{k=1}^{N'} \sum_{j=1}^{|\mathcal{F}|} m_j^k (\tilde{z}_j^k - z_j^k)^2 \tag{5}$$

where $m_j^k = 0$ if the ground truth forecast ($z_j^k$) is not available for the j$^{\text{th}}$ variable in the k$^{\text{th}}$ sample, and $m_j^k = 1$ if otherwise [9].

### 3.2.2. Pseudo-label extraction module

After pretraining the STraTS model by performing the forecast task, the model's last layer that is specific to the forecast task is removed, and then the resulting model is used to compute the non-temporal and time-series embeddings of the samples. The concatenation of the non-temporal embeddings and time-series embeddings of each sample is defined as its representation [9]. Then, K-means clustering analysis is performed on the learned representations. K-means takes the learned representations as input and divides them into $k$ subgroups using its geometric criterion [1]. It simultaneously learns a $d \times k$ centroid matrix $C$ and the cluster assignment $y_n$ of each subject $n$ as follows [20]:

$$\min_{C \in \mathbb{R}^{d \times k}} \frac{1}{N} \sum_{n=1}^{N} \min_{y_n \in \{0,1\}^k} \left\| f_\theta\left((d^n, T^n)\right) - C y_n \right\|_2^2 \text{ such that } y_n^\top 1_k = 1 \tag{6}$$

Optimizing this problem results in a set of optimal cluster assignments $(y_n^*)_{n \leq N}$ considered as the pseudo-labels of the subjects. That is, each subject $(d^n, T^n)$ is associated with a pseudo-label $y_n$ in $\{0,1\}^k$, representing the membership of the subject to one of the $k$ possible predefined classes.

### 3.2.3. Classification module

The extracted pseudo-labels are leveraged to supervise the training of a classifier $g_W$ predicting the accurate labels on top of the representations $f_\theta\left((d^n, T^n)\right)$ obtained from the STraTS model. Then, the classifier's parameters $W$ and the STraTS' parameters $\theta$ are simultaneously learned from the following optimization problem:

$$\min_{\theta, W} \frac{1}{N} \sum_{n=1}^{N} \ell\left(g_W\left(f_\theta((d^n, T^n))\right), y_n\right) \tag{7}$$

where $\ell$ is a negative log-softmax function.

SLAC-Time is an iterative procedure that alternates between two modules: (1) clustering the representations through the K-means algorithm and then using cluster assignments to generate pseudo-labels and (2) updating classifier's and STraTS model's parameters to predict the correct labels for each subject by minimizing the loss function.

## 4. Experimental results

### 4.1. Dataset

Our experiments were based on data obtained from the Transforming Research and Clinical Knowledge in Traumatic Brain Injury (TRACK-TBI) dataset [21]. This dataset, which was collected from 18 academic Level I trauma hospitals across the United States, includes detailed clinical data on 2996 TBI patients with different severity levels. The data used in this study includes 110 variables, including 59 non-temporal variables (demographics and one-time-recorded measurements at the time of emergency department (ED) visits) and 51 time-series variables collected during the first five days of TBI patients' hospital or ICU stay. We included three outcome variables, including the Glasgow Outcome Scale-Extended (GOSE) score, ICU length of stay, and mortality rate, to evaluate the validity of the identified TBI phenotypes. GOSE is a measure of functional outcome that assesses TBI patients in eight categories: (1) dead, (2) vegetative state, (3) lower severe disability, (4) upper severe disability, (5) lower moderate disability, (6) upper moderate disability, (7) lower good recovery, and (8) upper good recovery [22,23]. The ICU length of stay was defined as the duration of time that a patient spent in the ICU after admission, and the mortality rate was the proportion of patients who died within six months of the injury. We excluded the TBI patients with no GOSE score available. Considering this, 2160 TBI patients met the inclusion criteria. Non-temporal variables were not available for all the patients, so we performed iterative imputation to fill in the missing values in non-temporal variables. Both time-series and non-temporal variables were normalized to have zero mean and unit variance.

*Proxy task:* We perform time-series forecasting as a proxy task to pre-train the model and learn the initial representation of the multivariate time-series data. To do so, we define the set of observation windows as {24, 48, 72, 96, 118} hours and the prediction window as the 2-hour time period that comes just after the observation window. It should be noted that we only include the records with at least one time-series data in both observation and prediction windows. The data for performing time-series forecasting is divided into training and validation datasets with a ratio of 80 to 20.

*Target task:* The target task of SLAC-Time in this study is to subgroup TBI patients considering all the variables in the TRACK-TBI dataset that meet the inclusion criteria. These TBI patients are divided into training and validation datasets with a ratio of 80 to 20.

### 4.2. Implementation details

We implemented SLAC-Time using Keras and TensorFlow backend. **Table 1** represents the hyperparameters used in our experiments. Proxy task and target task models are trained using a batch size of 8 and Adam optimizer. Training for the proxy task is stopped when the validation loss does not decrease for ten epochs. Training for the target task is performed for 500 iterations, each of which is comprised of 200 epochs. Training in each iteration is also stopped when the validation loss does not decrease for ten epochs. The experiments were carried out on an NVIDIA Tesla P100 GPU, which took around four days.

Table 1. Hyperparameters used in the experiment

| Hyperparameter | Value |
| --- | --- |
| $M$ (Number of blocks in Transformer) | 2 |
| $d$ (Number of output neurons in FFNs) | 32 |
| $h$ (Number of attention heads in MHA) | 4 |
| Dropout | 0.2 |
| Learning rate | 0.0005 |

### 4.3. Optimal number of clusters

We evaluated how the quality of the clusters obtained by SLAC-Time is affected by the number of clusters, $k$. We performed the same proxy and downstream tasks using the same hyperparameters while changing $k$ according to the TBI clinicians' domain knowledge about the possible number of TBI phenotypes. We utilized various intrinsic clustering evaluation metrics, including the Silhouette coefficient, Calinski Harabasz index, Dunn index, and Davies Bouldin index on the data embeddings to measure the quality of the clusters.

Silhouette coefficient *s* for a sample is defined as

$$s = \frac{b - a}{\max(a, b)} \quad (8)$$

where *a* stands for the average distance between a sample in a cluster and the rest of the samples in the cluster, and *b* is the average distance between a sample in a cluster and the samples in the closest cluster. The Silhouette coefficient for the set of samples used in the clustering problem is given as the mean of the Silhouette coefficient for each sample. This score ranges from -1 for incorrect clustering to +1 for dense clustering. We used the Euclidean distance between embeddings as the distance metric. The choice of Euclidean distance is consistent with the distance metric used in the K-means clustering algorithm, which we employed for clustering the learned time-series representations in SLAC-Time.

Calinski Harabasz (CH) index is the ratio of dispersion between clusters to dispersion within clusters for all clusters defined as follows.

$$\left[\frac{\sum_{k=1}^{K} n_k \|c_k - c\|^2}{K - 1}\right] / \left[\frac{\sum_{k=1}^{K} \sum_{i=1}^{n_k} \|d_i - c_k\|^2}{N - K}\right] \quad (9)$$

Here, $n_k$ and $c_k$ are the number of samples and centroid of the $k^{th}$ cluster, respectively. c denotes the global centroid, and N is the total number of samples.

Dunn index is the ratio of the shortest distance between the samples from different clusters to the longest intra-cluster distance. The Dunn index varies from 0 to infinity with a higher index indicating higher quality of clusters. Dunn index of clustering with m clusters is represented as follows.

$$DI_m = \frac{\min \delta(C_i, C_j)}{\max \Delta_k} \quad (10)$$

$\delta(C_i, C_j)$ is the intra-cluster distance metric between clusters $C_i$ and $C_j$ where $1 \leq i \leq j \leq m$ and *m* stands for the total number of clusters. Also, $\Delta_k$ represents the maximum distance between observations in cluster *k* where $1 \leq k \leq m$.

Davies Bouldin index (DB) represents the average similarity between clusters where similarity is a metric that relates cluster distance to cluster size. This index is defined as

$$DB = \frac{1}{k}\sum_{i=1}^{k} \max R_{ij} \quad (11)$$

where $R_{ij}$ is the similarity between clusters i and j, and is calculated as follows.

$$R_{ij} = \frac{s_i + s_j}{d_{ij}} \quad (12)$$

where $s_i$ and $s_j$ are the intra-cluster dispersion of clusters i and j, respectively, and $d_{ij}$ represents the distance between the centroid of clusters.

We observed that k=3 results in the best clustering performance across all four clustering evaluation metrics (**Table 2**), suggesting three possible TBI phenotypes.

Table 2. SLAC-Time cluster quality for different numbers of clusters

| Number of clusters | Silhouette coefficient | Dunn index | Davies Bouldin index | Calinski Harabasz index |
|---|---|---|---|---|
| 3 | 0.06 | 0.09 | 1.9 | 90.3 |
| 4 | 0.05 | 0.08 | 5.3 | 54.3 |
| 5 | 0.03 | 0.06 | 7.8 | 34.6 |

**4.4. Number of cluster reassignments between iterations**

By updating the parameters of the model in each iteration, a record may be assigned to a new cluster. Accordingly, the cluster assignments may change over iterations. We measure the information shared between the clusters at iteration t-1 and t using the Normalized Mutual Information (NMI) defined as follows [24]:

$$\text{NMI}(A; B) = \frac{I(A; B)}{\sqrt{H(A)H(B)}} \quad (13)$$

where I is the mutual information and H denotes the entropy. If cluster assignments in iterations t-1 and t are perfectly dissimilar, the NMI will equal 0. If cluster assignments in iterations t-1 and t are perfectly the same, the NMI will equal 1. We measure NMI between the clusters at iterations t-1 and t to determine the actual stability of SLAC-Time. **Figure 4** demonstrates the NMI trend during 500 training iterations. As can be seen in **Figure 4**, the value of NMI significantly increases after about 200 iterations, indicating a decrease in cluster reassignments and an increase in the stability of clusters. NMI, however, stays below 1, meaning that several TBI patients are frequently reassigned between iterations.

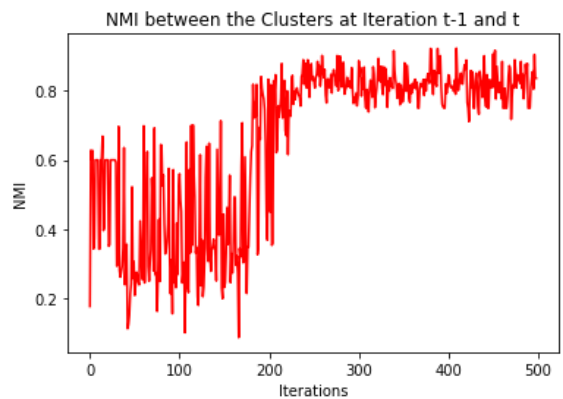

Figure 4. NMI trend during 500 training iterations

**4.5. Comparison of SLAC-Time and K-means clustering algorithm**

To demonstrate the effectiveness of SLAC-Time over the common clustering methods, we compared the clustering performance metrics of SLAC-Time with those of K-means for clustering multivariate time-series data in the TRACK-TBI dataset. Since there are no ground-truth labels regarding TBI phenotypes, we use different intrinsic clustering evaluation measures to quantify the clustering performance.

K-means is a common method for clustering both non-temporal and time-series data. In SLAC-Time, we use the K-means method to extract pseudo-labels of multivariate time-series data by clustering the learned representations. In order to cluster multivariate time-series data with missing values using K-means, it is necessary to handle missing values beforehand by imputing or interpolating them. This is because K-means requires complete data points for each variable to calculate distances and determine cluster membership. To enable a fair comparison between SLAC-Time and K-means, we handled missing values in the data using iterative imputation for non-temporal variables and linear interpolation for time-series variables. This allowed us to create complete data points for both types of variables and ensure an unbiased evaluation of the performance of both clustering algorithms. The clustering evaluation metrics show that SLAC-Time outperforms K-means (**Table 3**), suggesting that clustering multivariate time-series data based on the learned representations rather than raw data can be more effective.

Table 3. Comparison of SLAC-Time and K-means clustering method based on different clustering quality metrics

| Clustering method | Silhouette coefficient | Dunn index | Davies Bouldin index | Calinski Harabasz index |
| --- | --- | --- | --- | --- |
| K-means | 0.07 | 0.11 | 4.31 | 131.66 |
| SLAC-Time | 0.13 | 0.15 | 1.8 | 196.31 |

### 4.6. Characteristics of the TBI phenotypes

To discover distinct characteristics of TBI phenotypes, we examine the variables that have been shown to be explanatory in the prediction of the 6-month GOSE score [25,26], and those that are significant based on TBI clinicians' domain knowledge. These variables include age, sex, overall GCS score, GCS motor score, GCS eye score, pupil reactivity, hypoxia, hypotension, intubation, glucose, hemoglobin, white blood cell (WBC), hematocrit, international normalized ratio (INR) and activated partial thromboplastin time (aPTT). For making comparisons between the phenotypes, non-temporal variables were represented by mean and standard deviation or numbers and percentages (**Table 4**). We also represent the phenotype-specific average of time-series variables with 95% confidence intervals and the correlation between them (**Figures** 6-13). We univariately analyze the important non-temporal and time-series variables and test whether there is a significant difference between the phenotypes using the Kruskal-Wallis test. The differences between phenotypes were considered significant when corresponding p-values are <0.05.

Using k=3 (the optimal number of clusters) as the number of TBI phenotypes, we performed SLAC-Time for clustering 2160 TBI patients in the TRACK-TBI dataset, which resulted in three TBI phenotypes $\alpha$, $\beta$, and $\gamma$ that include 693, 586, and 881 TBI patients, respectively. Each TBI phenotype had a distinct baseline feature profile linked to its outcome endpoints.

Table 4. Key demographics and non-temporal clinical features of TBI patients included in clustering analysis

| Feature | Phenotype α | Phenotype β | Phenotype γ |
| --- | --- | --- | --- |
| Total subjects, n | 693 | 586 | 881 |
| Age | | | |
|     Age, mean ± SD | 28 ± 14 | 41 ± 18 | 48 ± 18 |
|     Age≤30, n (%) | 471 (68%) | 223 (38%) | 182 (21%) |
|     30<Age≤45, n (%) | 139 (20%) | 123 (21%) | 217 (24%) |
|     45<Age≤60, n (%) | 69 (10%) | 142 (24%) | 245 (28%) |
|     Age>60, n (%) | 14 (2%) | 98 (17%) | 237 (27%) |
| Sex | | | |
|     Male, n (%) | 336 (53%) | 426 (72%) | 720 (81%) |
|     Female, n (%) | 357 (47%) | 160 (28%) | 161 (19%) |
| Clinical variables | | | |
| ED Glucose, mean ± SD | 122 ± 32 | 158 ± 67 | 138 ± 58 |
| ED Hemoglobin, mean ± SD | 13.7 ± 1.6 | 13.5 ± 1.8 | 14.2 ± 1.6 |
| ED INR, mean ± SD | 1.06 ± 0.09 | 1.22 ± 0.86 | 1.07 ± 0.17 |
| Hypoxia, n (%) | 19 (2.7%) | 37 (6.3%) | 13 (1.5%) |
| Hypotension, n (%) | 17 (2.5%) | 51 (8.7%) | 14 (1.6 %) |
| GCS score | | | |
| ED GCS score, mean ± SD | 14 .1 ± 2.1 | 8.5 ± 5.1 | 13.9 ± 2.7 |
| ED GCS motor score | | | |
| 1 (no response), n (%) | 4 (0.6%) | 16 (2.7%) | 3 (0.3%) |
| 2 (extension), n (%) | 5 (0.7%) | 12 (2%) | 4 (0.5%) |
| 3 (flexion abnormal), n (%) | 21 (3%) | 46 (8%) | 8 (1%) |
| 4 (flexion withdrawal), n (%) | 18 (2.6%) | 85 (14.5%) | 27 (3%) |
| 5 (localizes to pain), n (%) | 581 (84%) | 193 (33%) | 783 (89%) |
| 6 (obeys commands), n (%) | 2 (0.3%) | 33 (5.6%) | 5 (0.6%) |
| ED Pupil reactivity | | | |
|     Both pupils react, n (%) | 603 (87%) | 349 (60%) | 730 (83%) |
|     Only one pupil reacts, n (%) | 0 (0%) | 22 (3.8%) | 8 (1%) |
|     Neither of the pupils reacts, n (%) | 1 (0.14%) | 96 (16.38%) | 3 (0.34%) |

*Comparison of Outcomes across TBI Phenotypes:* Evaluating the outcome variables of the TBI patients in each phenotype showed that phenotypes α, β, and γ significantly differ from one another. Phenotypes α and β had the best and worst across all three outcomes, respectively (**Figure 5**). Phenotype α had the best GOSE score (6.9 ± 1) among TBI phenotypes. 68% of TBI patients in phenotype α had a GOSE score of 7 or 8, while 27% and 47% of the patients in phenotypes β and γ had a GOSE score of 7 or 8, respectively. Phenotype α is also characterized by the lowest mortality rate (0.6%) and the shortest ICU length of stay (4 days) compared to the other two phenotypes. Phenotype β with the lowest GOSE score (5.1 ± 2.2) had the worst recovery compared to the other two phenotypes. 15% of the TBI patients in phenotype β had a GOSE score of 1 or 2, while only 1% and 4% of the TBI patients in phenotype α and γ have such GOSE scores, respectively. Besides having the lowest GOSE score, phenotype β is characterized by the highest mortality rate (14%) and longest ICU stay (21 days). Even though phenotype γ with an average GOSE score of 6.5 ± 1.5 seems to overlap with phenotype α, these two phenotypes are significantly different in terms of both GOSE score (p < 0.05) and ICU length of stay (p < 0.05). Furthermore, the higher mortality rate of phenotype γ (3.5%) compared to that of phenotype α (0.6%) emphasizes the higher severity of phenotype γ compared to phenotype α.

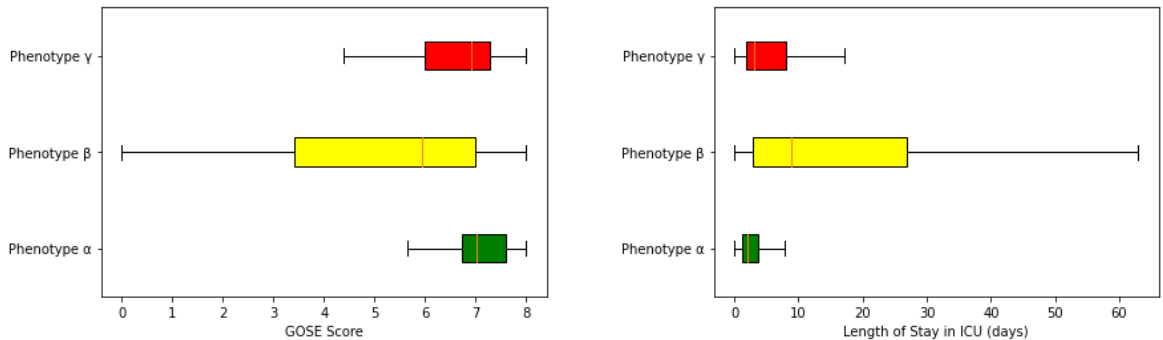

Figure 5. Boxplot illustrating GOSE scores and ICU lengths of stay for TBI patients across different phenotypes

*Comparison of Sex and Age across TBI Phenotypes*: Phenotype α with an average age of 28 ± 14 was the youngest group (p < 0.005). 68% of the TBI patients in phenotype α were 30 years old or younger, and only 2% of the patients in this phenotype were older than 60. On the other hand, phenotype γ had the oldest group (48 ± 18). 21% of TBI patients in phenotype γ were younger than 30, and 27% were older than 60. Most of the TBI patients were male (**Table 4**), and there is a significant difference between TBI phenotypes in terms of their gender (p < 0.005). Phenotype α had the highest percentage of females (47%) among all phenotypes, whereas only 19% of TBI patients in phenotype γ were female.

*Comparison of ICU Admission Rate across TBI Phenotypes*: Of the TBI patients in the TRACK-TBI dataset, less than 1% are discharged from the ED, while the vast majority require admission to hospitals for ongoing treatment and management following ED care. Those admitted to hospitals are transferred to ICUs if they suffer from a serious injury and need special care. About 90% of TBI patients in phenotype β were transferred to ICUs, which suggests that the patients in this phenotype primarily had severe and life-threatening injuries requiring intensive care. Also, 30% and 48% of TBI patients in phenotype α and phenotype γ were transferred to ICUs, respectively, meaning that the TBI patients in phenotype γ had a more severe injury than those in phenotype α, and are more likely to need intensive care after hospital admission.

*Comparison of Level of Consciousness and Pupil Reactivity across TBI Phenotypes:* Higher severity of phenotype β compared to phenotypes α and γ is evident in their GCS motor scores, GCS eye scores, and pupil reactivity (**Figures 6-8**). More than 40% of the TBI patients in phenotype β had no motor response once they are admitted to the ICU, while about 4% and 10% of TBI patients in phenotype α and γ had no motor response upon ICU admission, respectively (**Figure 6**). Likewise, about 60% of the TBI patients in phenotype β have no eye-opening response once they are admitted to the ICU, while only 8% and 20% of the TBI patients in phenotypes α and γ have no eye-opening response upon ICU admission, respectively (**Figure 7**). For all three phenotypes, the percentage of TBI patients with no motor response and no eye-opening response sharply decreases on the first day of ICU stay. The highest GCS motor score (score 6: obey commands) and the highest GCS eye score (score 4: response spontaneously) were significantly different among the phenotypes ($p < 0.001$). Phenotype β had the lowest percentage of TBI patients with GCS motor score equal to 6. Only about 24% of the TBI patients in phenotype β had the highest GCS motor and eye scores once they are admitted to the ICU. Although the percentages of TBI patients in phenotype β with the highest GCS scores increases over time, they remain much lower than those in phenotypes α and γ during the ICU stay. Furthermore, phenotypes α and γ differ significantly in all the aforementioned GCS motor and GCS eye score categories ($p < 0.001$). Phenotype α had the lowest percentages of TBI patients with no GCS motor or no GCS eye response. On the other hand, it had the highest percentages of TBI patients with the best GCS motor and eye scores.

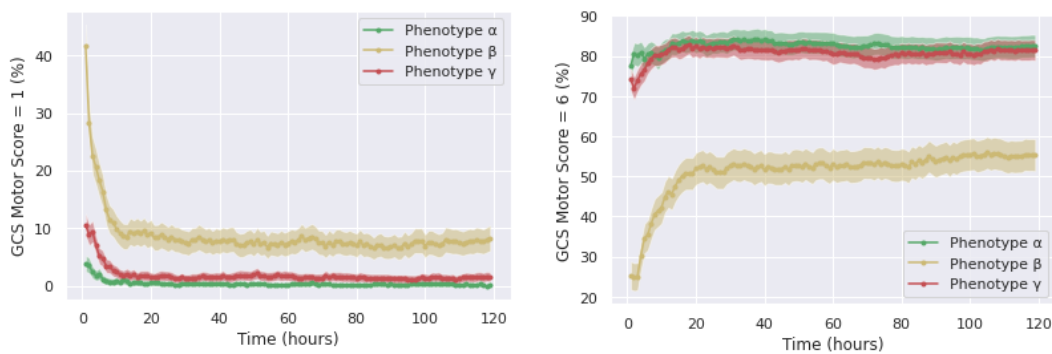

Figure 6. Percentage of TBI patients with the lowest and highest GCS motor during the first 120 h of ICU stay

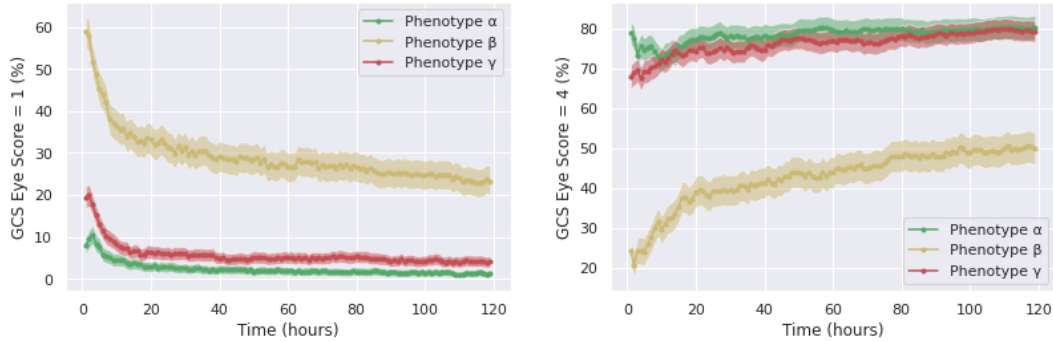

Figure 7. Percentage of TBI patients with the lowest and highest GCS eye score during the first 120 h of ICU stay

Pupil reactivity is among the variables with the highest contribution to the long-term recovery outcome of TBI patients [25,26]. There are significant differences between the pupil reactivity of TBI phenotypes in both ED and the ICU ($p < 0.001$). Phenotype $\alpha$ and phenotype $\gamma$ differ significantly in terms of pupil reactivity ($p < 0.001$). The percentages of TBI patients with no pupil reactivity or with only one reactive pupil are less in phenotype $\alpha$ compared to phenotype $\gamma$. During ED visits, phenotype $\alpha$ had the highest percentage of TBI patients with two reactive pupils (87%) (**Table 4**). On the other hand, during ICU stay, the percentage of phenotype $\alpha$ patients with two reactive pupils is slightly less than that of phenotype $\gamma$ (**Figure 8**). This difference can be due to the missing values in the time-series variable associated with pupil reactivity. TBI patients in phenotype $\beta$ had the worst pupil reactivity among the TBI patients in both ED and ICU. Phenotype $\beta$ had the highest percentage of TBI patients with no pupil reactivity and the lowest percentage of TBI patients, both of whose pupils react.

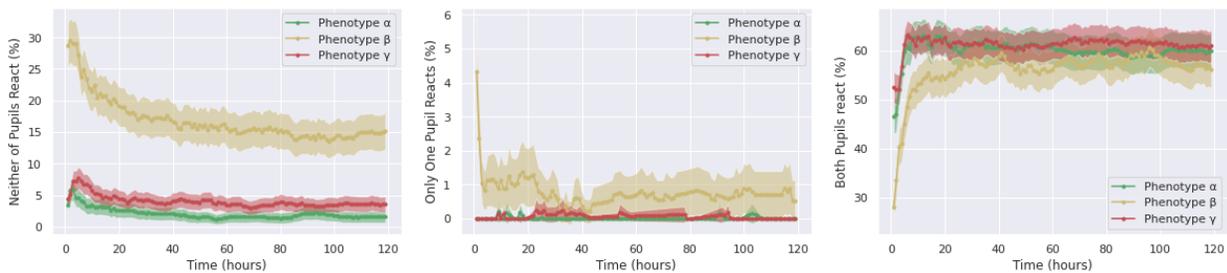

Figure 8. Percentage of TBI patients in each category of pupil reactivity during the first 120 h of ICU stay

*Comparison of Clinical Variables across TBI Phenotypes*: The analysis of phenotype-specific averages of clinical variables reveals significant differences among the three TBI phenotypes. Phenotype β exhibited the highest rates of hypoxia and hypotension compared to the other two phenotypes ($p < 0.005$). Additionally, the rates of hypoxia and hypotension in phenotype α were significantly higher than those in phenotype γ ($p < 0.005$) (**Table 4**).

In **Figure 9**, we present the average hematocrit, hemoglobin, and WBC values for each TBI phenotype during the first 120 hours of ICU stay. Phenotype β had the lowest hemoglobin and hematocrit values as well as the highest WBC count, indicating more severe blood loss and injury compared to the other two phenotypes. This finding is consistent with the lowest GCS score in the ED (8.5 ± 5.1) and the worst GCS motor response, GCS eye response, and pupil reactivity trajectories during the five-day ICU stay (**Figures 6-8**).

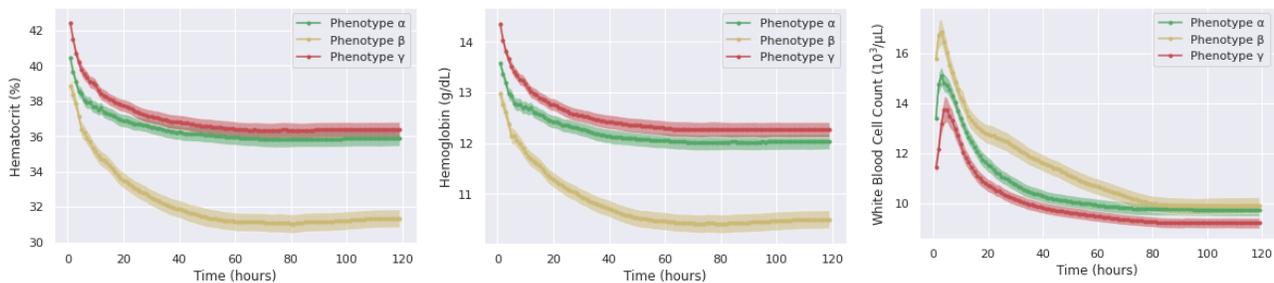

Figure 9. Mean hematocrit, hemoglobin, and WBC levels for TBI patients in each phenotype during the first 120 h of ICU stay

In addition, phenotype β had the highest intubation rate. About 60% of the TBI patients in phenotype β are intubated once they are admitted to the ICU, while phenotype α had the lowest intubation rate and only 10% of the TBI patients in this phenotype were intubated upon ICU admission (**Figure 10**). Likewise, phenotype β and phenotype α had the highest and lowest levels of glucose levels during ICU stay, accordingly (**Figure 11**).

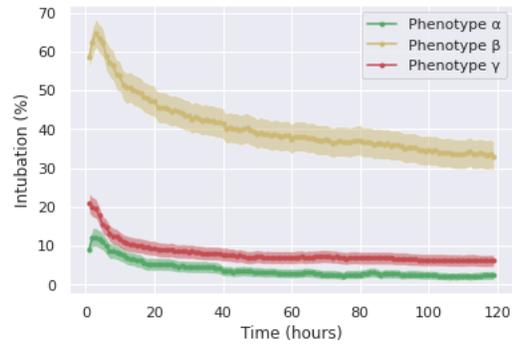

Figure 10. Percentage of TBI patients intubated in each phenotype during the first 120 h of ICU stay

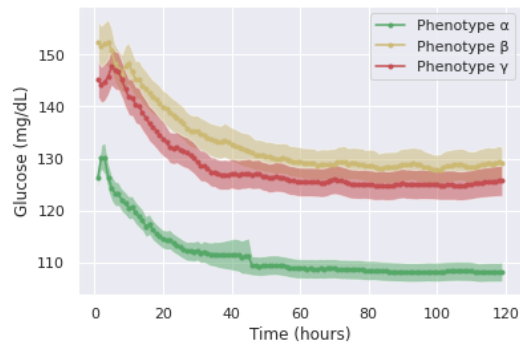

Figure 11. Mean glucose levels for TBI patients in each phenotype during the first 120 h of ICU stay

The INR and aPTT tests are used to measure how quickly blood clots in different pathways [27]. As can be seen in **Figure 12**, phenotype β had the highest levels of aPTT and INR among the TBI phenotypes, suggesting that phenotype β includes patients suffering from bleeding disorders due to their TBI. On the other hand, the relatively normal coagulation measurements of phenotypes α and γ align with their low bleeding rates.

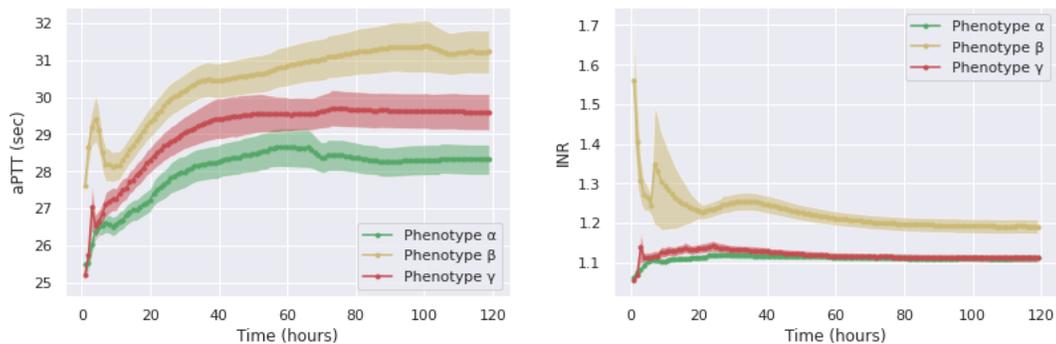

Figure 12. Mean aPTT and INR levels for TBI patients in each phenotype during the first 120 h of ICU stay

*Comparison of Heterogeneity and Dispersion across TBI Phenotypes*: **Figure 13** presents a Principal Component Analysis (PCA) plot of TBI patients, emphasizing the high heterogeneity of TBI patients, as there is no clear distinction between them in the PCA plot, particularly between phenotypes α and γ. The plot shows that phenotype β has a substantially larger dispersion than phenotypes α and γ, aligning with our previous findings that indicate worse clinical outcomes for phenotype β, such as the lowest GOSE scores, highest mortality rates, and longest ICU stays. The PCA plot reveals that patients in phenotypes α and γ are located near each other, with phenotype γ exhibiting greater dispersion than phenotype α. This difference in dispersion highlights the varying levels of heterogeneity within each group, suggesting that the broader range of underlying pathophysiological features in phenotype γ could contribute to its worse clinical outcomes compared to phenotype α.

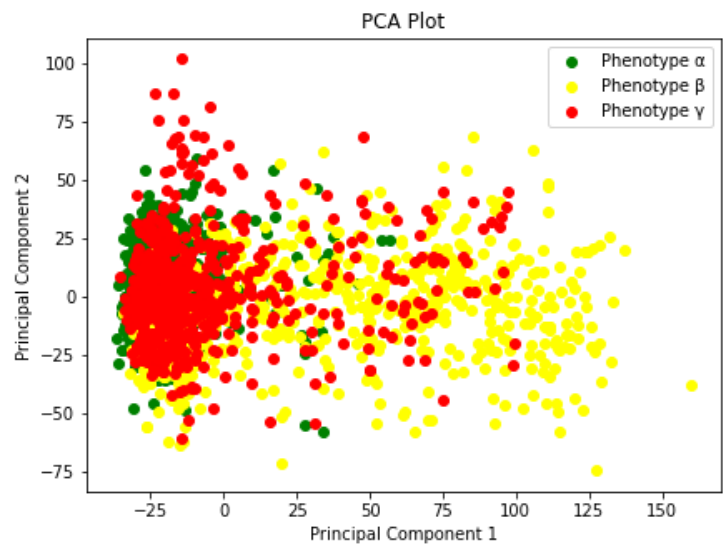

Figure 13. PCA plot of TBI Patients in each phenotype based on two principal components

*Comparison of Correlations between Important Variables across TBI Phenotypes*: We found that glucose, hematocrit, hemoglobin, and WBC values are negatively correlated with the best GCS motor response (obey commands), best GCS eye response (response spontaneously), and best pupil reactivity (both pupils react) across all three phenotypes. Conversely, these variables are positively correlated with the worst GCS motor response (no response), worst GCS eye response (no response), and worst pupil reactivity (neither pupil reacts) across all three phenotypes (see **Figure 14**). In other words, a decrease in these four time-series variables during a TBI patient's ICU stay may indicate the recovery of their impaired consciousness.

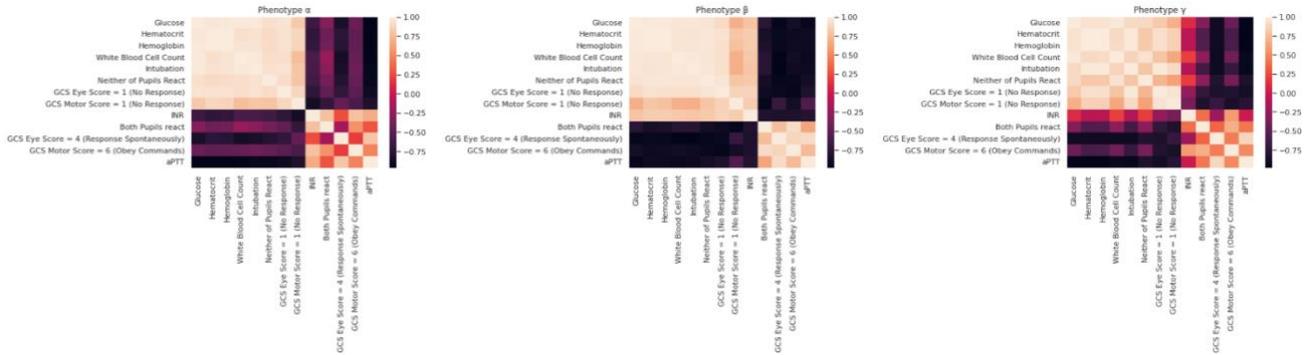

Figure 14. Correlation analysis of time-series features within each TBI phenotype

## 5. Discussion

We applied SLAC-time to the clustering of TBI patients based on non-temporal and time-series clinical variables during the first five days of ICU stay. We identified three TBI phenotypes ($\alpha$, $\beta$, and $\gamma$) that are distinct from one another in terms of clinical variables and outcome endpoints. Phenotype $\beta$ had the worst clinical outcomes. Phenotype $\alpha$ had better clinical outcomes than those in phenotype $\gamma$, although phenotype $\alpha$ had less favorable trends than phenotype $\gamma$ in some clinical variables such as WBC, hematocrit, and hemoglobin as well as hypoxia, and hypotension rates.

WBCs circulate through the bloodstream and tissues to respond to the injury and protect the body from infection after trauma [28]. Considering this, the more severe the injury, the higher the WBC counts [29]. Hematocrit is the percentage of red blood cells in the blood, and hemoglobin enables red blood cells to carry oxygen and $CO_2$ throughout the body. After trauma, low hemoglobin or hematocrit levels indicate that the patient is losing red blood cells because of acute bleeding. Comparing the values of WBC, hemoglobin, and hematocrit in TBI phenotypes demonstrates that phenotype $\beta$ had the highest blood loss among the TBI phenotypes. Furthermore, phenotype $\alpha$ suffers from more blood loss compared to phenotype $\gamma$. The amount and intensity of blood loss in TBI phenotypes can also be seen in their rates of hypoxia and hypotension. Hypoxia is the absence of oxygen in the body tissues, and it can be caused due to a low number of blood cells from severe blood loss. Hypotension is defined as having a blood pressure of less than 90/60 mm/Hg i.e., low blood pressure. Low blood volume due to severe blood loss can cause low blood pressure. The rates of hypoxia and hypotension substantiate that phenotype β had the highest and phenotype γ had the lowest blood loss.

Phenotype α, despite having more blood loss compared to phenotype γ, had better recovery outcomes. This might be because the patients in phenotype α are much younger than phenotype γ. This supports the cluster analysis because it is consistent with the assumption that younger age yields better TBI outcomes, especially for more severe cases. Gender might be another reason why phenotype α has better outcomes compared to phenotype γ. Phenotype α with a higher percentage of females result in a better outcome. This aligns with clinical research that suggests better outcomes and recovery of females compared to males after injury [30,31]. The glucose levels might also contribute to the slower recovery of phenotype γ compared to phenotype α. Over time, a high glucose level, indicative of a high blood sugar level, might harm the body's organs and result in potential long-term effects by damaging small and big blood vessels [32]. Clinical studies shows that high blood sugar levels harm the brain by raising intracranial pressure, and it contributes to poorer outcomes after injury [33]. An uncontrolled glucose level may impede or delay the recovery of TBI patients. The higher glucose levels of phenotype γ might explain why the patients in this phenotype recover more slowly than those of phenotype α. All three phenotypes had their highest level of glucose once they are admitted to the ICU, which is associated with disturbed cerebrovascular pressure reactivity following TBI [33].

The potential applicability of SLAC-Time extends beyond the realm of TBI to other clinical clustering tasks in areas such as infectious diseases, cardiology, oncology, and chronic disease management. For instance, it can enable the clustering of patients with infectious diseases such as COVID-19 based on symptom progression and treatment responses, helping clinicians identify patient subgroups and tailor personalized interventions. SLAC-Time is also well-suited for clustering time-series data in other domains such as finance, energy management, and environmental monitoring applications. In each of these domains, time-series data often presents unique challenges, including complexity, non-stationarity, and missing values—similar to those observed in TBI datasets. By successfully addressing these challenges in the context of TBI data, we demonstrate the potential of SLAC-Time to be readily adapted and employed in these diverse fields. The self-supervised learning and the ability to handle missing values without imputation or aggregation make it a valuable tool for a broad range of applications where accurate and efficient clustering of time-series data is essential.

Finally, we chose to use the K-means method for clustering time-series representations within SLAC-Time for its simplicity, efficiency, and widespread use in the field of time-series analysis. However, SLAC-Time is a flexible framework that can accommodate a variety of clustering algorithms. The choice of clustering algorithm should depend on the specific characteristics of the dataset and the problem domain. For instance, density-based clustering algorithms such as DBSCAN may be more appropriate when the data exhibits clusters of varying densities or when noise is present. By incorporating alternative clustering techniques, SLAC-Time can be further customized to address a broader range of problems. It is essential for researchers to consider factors such as the underlying distribution of the data, the presence of noise, and the computational complexity of the algorithm when selecting the most appropriate clustering technique for their specific application.

**5.1. Limitations**

There are several limitations to this study. First, we applied SLAC-Time to a TBI-specific dataset with longitudinal outcomes. However, we may need to evaluate SLAC-Time using a broader clinical multivariate time-series dataset. Second, our study was primarily focused on TBI patients within the TRACK-TBI dataset, and we acknowledge that one limitation is our inability to externally validate the derived phenotypes. This constraint mainly stems from the scarcity of alternative datasets that possess the requisite temporal data and clinical outcomes necessary for effective validation. In addition, challenges in accessing and obtaining approvals for using such datasets further contribute to this limitation. Third, our study was inevitably limited by the available clinical variables. More insights into phenotypical differences can be derived if additional clinical variables such as specific CT results, neurological symptoms, and genetic profiles of the TBI patients had been used in clustering. Finally, SLAC-Time is specifically designed to handle missing values in multivariate time-series data, but not in non-temporal data. In cases where non-temporal data contains missing values, one approach is to represent non-temporal variables as triplets, with a default time value. However, this method may affect the model's performance [9]. Therefore, it may be necessary to use data imputation methods to fill in the missing values in non-temporal data.

## 6. Conclusion

We proposed a self-supervised learning-based approach to cluster multivariate time-series data with missing values without resorting to data imputation and aggregation methods. We used time-series forecasting as a proxy task to learn the representation of unlabeled multivariate time-series data. SLAC-Time iteratively clusters the representations with K-means and updates its parameters by predicting cluster labels as pseudo-labels. SLAC-Time outperforms K-means in all clustering evaluation metrics, suggesting that using learned representations rather than raw data for clustering multivariate time-series data might mitigate the negative influence of noises in raw data on clustering performance. SLAC-Time needs limited to no domain knowledge about input data, making it an excellent choice for clustering multivariate time-series data in fields where data annotations are rare. The performance of SLAC-Time for clustering TBI patients demonstrates its applicability to sparse and irregularly sampled multivariate time-series data. We successfully derived TBI phenotypes ($\alpha$, $\beta$, and $\gamma$) from the TRACK-TBI dataset, revealing significant differences between their outcomes. The identification of these distinct TBI phenotypes has significant implications for designing clinical trials and developing treatment strategies tailored to the specific physiological characteristics of TBI patients. By considering the unique features and needs of each TBI phenotype, researchers and clinicians can develop more targeted interventions with the potential to improve outcomes and reduce the likelihood of ineffective or harmful interventions. Additionally, a nuanced understanding of TBI phenotypes can inform the development of new diagnostic tools and treatment approaches designed to address the underlying mechanisms of each subtype. We recommend that future research focus on external validation by examining cohorts from other TBI datasets to ensure the generalizability of the phenotypes. Additionally, exploring alternative self-supervision tasks for the STraTS model and developing a framework for interpreting self-supervised learning-based clustering of multivariate time-series data are recommended avenues for further research.

**Acknowledgments**

This material is based upon work supported by the National Science Foundation under grants #1838730 and #1838745. Dr. Foreman was supported by the National Institute of Neurological Disorders and Stroke of the National Institutes of Health (K23NS101123). The content is solely the responsibility of the authors. Any opinions, findings, and conclusions or recommendations expressed in this material are those of the authors and do not necessarily reflect the views of the National Science Foundation or of the National Institutes of Health. The authors acknowledge the TRACK-TBI Study Investigators for providing access to data used in this work.